\documentclass[journal]{IEEEtran}

\usepackage[utf8]{inputenc}
\usepackage{authblk}
\usepackage{graphicx}
\usepackage{cite}
\usepackage{indentfirst}
\usepackage[justification=centering]{caption}
\usepackage{amsmath}
\usepackage{lscape}
\usepackage{gensymb}
\usepackage{algorithm}
\usepackage{algpseudocode}
\usepackage{mathtools}
\usepackage{amsfonts}
\usepackage{comment}
\usepackage{bm}
\algnewcommand{\Inputs}[1]{%
  \State \textbf{Inputs:}
  \Statex \hspace*{\algorithmicindent}\parbox[t]{.8\linewidth}{\raggedright #1}
}
\algnewcommand{\Initialize}[1]{%
  \State \textbf{Initialize:}
  \Statex \hspace*{\algorithmicindent}\parbox[t]{.8\linewidth}{\raggedright #1}
}
\usepackage{hyperref}
\usepackage[dvipsnames]{xcolor}
\DeclareMathOperator*{\argmin}{arg\,min}  

\graphicspath{ {Images/} }

\begin{document}

\title{A Neuromorphic Vision-Based Measurement for Robust Relative Localization in Future Space Exploration Missions}

\author{Mohammed~Salah,
        Mohammed~Chehadah,~\IEEEmembership{Member,~IEEE,}
        Muhammad~Humais,
        Mohammed~Wahbah,
        Abdulla Ayyad,~\IEEEmembership{Member,~IEEE,}
        Rana Azzam,
        Lakmal~Seneviratne,
        and~Yahya~Zweiri,~\IEEEmembership{Member,~IEEE}
\thanks{This work was supported by  the Khalifa University of Science
and Technology Award No. RC1-2018-KUCARS. Mohammed Salah is the corresponding author (email: 100058291@ku.ac.ae)}
\thanks{M. Salah, M. Chehadah, M. Humais, M. Wahbah, R. Azzam, L. Senevirante, and Y. Zweiri are with the Center for Autonomous Robotic Systems, Khalifa University, Abu Dhabi, United Arab Emirates. A. Ayyad is with Advanced Research and Innovation Center (ARIC), Khalifa University of Science and Technology, Abu Dhabi, United Arab Emirates. Also, L. Seneviratne is the director of Center for Autonomous Robotic Systems, and Y. Zweiri is with the Department of Aerospace Engineering, both at Khalifa University, Abu Dhabi, United Arab Emirates.}}

\maketitle

\begin{abstract}
Space exploration has witnessed revolutionary changes upon landing of the Perseverance Rover on the Martian surface and demonstrating the first flight beyond Earth by the Mars helicopter, Ingenuity. During their mission on Mars, Perseverance Rover and Ingenuity collaboratively explore the Martian surface, where Ingenuity scouts terrain information for rover's safe traversability. Hence, determining the relative poses between both the platforms is of paramount importance for the success of this mission. Driven by this necessity, this work proposes a robust relative localization system based on a fusion of neuromorphic vision-based measurements (NVBMs) and inertial measurements. The emergence of neuromorphic vision triggered a paradigm shift in the computer vision community, due to its unique working principle delineated with asynchronous events triggered by variations of light intensities occurring in the scene. This implies that observations cannot be acquired in static scenes due to illumination invariance. To circumvent this limitation, high frequency active landmarks are inserted in the scene to guarantee consistent event firing. These landmarks are adopted as salient features to facilitate relative localization. A novel event-based landmark identification algorithm using Gaussian Mixture Models (GMM) is developed for matching the landmarks correspondences formulating our NVBMs. The NVBMs are fused with inertial measurements in proposed state estimators, landmark tracking Kalman filter (LTKF) and translation decoupled Kalman filter (TDKF) for landmark tracking and relative localization, respectively. The proposed system was tested in a variety of experiments and has outperformed state-of-the-art approaches in accuracy and range.
\end{abstract}


\IEEEpeerreviewmaketitle

\section{Introduction}
\label{section:intro}
\IEEEPARstart{S}{pace} exploration has delivered a wealth of scientific discoveries on the history and evolution of the solar system. Space exploration recently witnessed a paradigm shift as space agencies started to deploy unmanned platforms for various space missions allowing exploration of areas that were once inaccessible to mankind. NASA marked a major milestone deploying the first rover-copter team comprised of the Mars helicopter, Ingenuity, and the Mars rover, Perseverance, on Mars \cite{copter_demonstrator}. The Ingenuity scouts the terrain to facilitate terrain mapping and boost the rover's navigation performance. However, relative localization between both platforms is of fundamental importance for their safe navigation.

 Traditional relative localization systems relying on radio-frequency identification (RFID) and ultra-wideband (UWB) do not provide sufficient relative localization accuracy \cite{uwb_vision_survey,uwb_vision_survey_2}. On the other hand, vision-based measurements (VBMs) using conventional imaging sensors tend to deteriorate in harsh space environments \cite{scramuzza_space}. To alleviate the shortcomings of such systems such as low dynamic range and high latency, there is a surging demand for new relative localization paradigms based on robust sensing modalities that could endure harsh environmental conditions and hence enable persistent relative localization.

Several approaches have been proposed for relative localization systems, mainly relying on radio frequency identification (RFID), ultrawideband (UWB) measurements, VBM, and neuromorphic vision-based measurements (NVBM). RFID-based measurements have been generally utilized for indoor localization systems due to their passive nature \cite{rfid_survey_1,rfid_survey_2,rfid_1,rfid_2}. It was not until recently, though, that RFID-based relative localization systems were developed due to the large tracking errors of onboard antennae. Li et al. \cite{rfid_1} proposed a relative localization system that relies on phase, received signal strength indicator (RSSI), and readability that is tolerant to the antenna tracking errors. However, such relative localization systems are impractical for space applications since RFID measurements require a fixed infrastructure for the tags.  

UWB-based approaches are considered as one of the earliest approaches to relative localization, due to the low cost and ease of deployment of UWB. Previously, UWB measurements were utilized for localization in GPS-denied environments where fixed nodes were installed as anchors for positioning the target platform \cite{uwb_survey_1,uwb_survey_2}. Similar to RFID, such localization systems are difficult to utilize for space applications for the fixed infrastructure requirement. To overcome this challenge, Strader et al. \cite{strader_uwb} proposed a relative localization system that estimates the relative pose between two UAVs in-motion, augmenting range and displacement measurements from UWB sensor and a generic navigation system using nonlinear least squares. However, the approach was only tested in simulation and needs experimentation to validate robustness. Guler et  al. \cite{guler_uwb} proposed a different approach based on Monte Carlo localization (MCL) sequential methods for modelling the motion of the tag quadrotor. The quadrotor's motion was restricted to the horizantal plane where it was flying at constant altitude.

Landmark-based positioning systems proved to be superior to UWB and RFID in terms of localization accuracy. In such systems, conventional imaging sensors extract landmark positions as VBM to reconstruct the agent pose \cite{yusra_survey,indoorvlc,vlc_ofdm,indoor_survey}. Walter et al. \cite{walter_markers} proposed a relative localization system for quadrotors relying on a conventional camera with ultraviolet pass filter and ultraviolet LEDs emitting light at frequencies much lower than the visible light spectrum. Similarly, Cutler et al. \cite{cutler_markers} utilized an infrared camera with infrared emitting landmarks. However, the major drawback of these systems is the degraded performance of the conventional imaging sensors under extreme space conditions due to their high latency and low dynamic range. Hence, utilizing a different imaging pipeline is substantial for the robustness of such systems in uncertain space environments.

Neuromorphic vision sensors (NVS) provide improved performance in extreme environments given their high dynamic range and low latency achieved by virtue of their asynchronous nature \cite{event_survey, tobi_dvs, vision_transformer}, which motivate the move towards neuromorphic vision technologies. These unprecedented capabilities of the NVS have been utilized in space \cite{scramuzza_space, star_tracking}, tactile sensing \cite{tactile_sensors, yahya_tactile, yahya_tactile_2}, and autonomous navigation applications \cite{event_odometry_1, event_odometry_2}. More importantly, the NVS is currently utilized as an imaging sensor for various vision-based localization systems. Mueggler et al. \cite{6dof_pose} and Valeiras et al. \cite{3d_pose_estimation} show the superiority of the NVS over conventional imaging sensors for UAV navigation as traditional cameras suffer from motion blur and degraded feature extraction from the observed scenes. Nguyen et al. \cite{event_lstm} proposed a stacked spatial LSTM network to robustly extract the features observed from accumulated event frames and learn their spatial dependencies in the image feature space. Yuan and Ramalingam \cite{fast_nvs} developed a fast spatio-temporal binning technique to detect lines that correspond to edges in the observed scene as features for localization. In all of these approaches, determining the relative motion between the NVS and the observed scene is a must for events to be generated, posing a major limitation in absence of activity in the environment. To alleviate this restriction, flickering LED luminaries are used as a constant source of intensity variations to ensure consistent event generation and as features for localization \cite{flicker2,flicker,grasping, pose_tracking}, irrespective of the scene dynamics. Chen et al. \cite{flicker2} proposed an indoor localization system using an NVS and flickering landmarks where the landmarks are characterized by unique frequency IDs without introducing complex data association algorithms. The drawback of these systems is the high levels of noise present in frequency detections leading to false landmarks identifications and large positioning errors. In addition, due to the sensor's low resolution, the positioning accuracy of these systems, 0.03 m within 1 meter range, which is insufficient for the targeted space scenario.

Previously developed relative localization systems suffer in terms of installation feasibility in space and their robustness in extreme environments. On the other hand, neuromorphic vision along with flickering landmarks demonstrates an improved system architecture for landmark-based positioning systems. However, robust landmark identification and localization accuracy remain a gap that needs to be addressed accordingly. 

NVSs have a huge potential for future space missions given their unprecedented capabilities in harsh environmental conditions. Current space applications of the NVS are mainly limited to astronomical object tracking, such as stars and space debris \cite{star_tracking,space_debris} and visual odometry for planetary robots \cite{scramuzza_space}. In this work, we use NVS to enable relative localization under harsh space conditions. In our approach, neurmorphic vision-based measurements (NVBM) are fused with inertial measurements in robust state estimators providing state-of-the-art (SOTA) relative localization performance.

\subsection{Contributions} \label{subsection:contributions}
The motivation of this article is to develop a reliable relative localization system for the Ingenuity and Perseverance Rover providing robust landmark identification and relative localization performance. Accordingly, this article comprises the following contributions:

\begin{enumerate}
    \item A novel event-based landmark identification system is proposed to detect and identify salient flickering features by means of a frequency-based unsupervised clustering using Gaussian Mixture Models (GMMs) with a maximum frequency detection error of 3.21 Hz.
    \item A measurement fusion of NVBMs and inertial measurements is developed using a landmark tracking Kalman filter (LTKF) and a translation decoupled Kalman filter (TDKF) for landmark tracking and relative localization.
    \item The developed relative localization system was demonstrated through a space exploration scenario outperforming state-of-the-art approaches in terms of accuracy with a mean positioning error of 0.0052 m.
\end{enumerate}

\subsection{Structure of The Article} \label{subsection:structure}
The rest of the article is organized as follows. Section \ref{section:overview} provides the notation followed in the paper and gives an overview of the proposed relative localization system. Section \ref{section:LTKF} discusses the theory behind the NVBMs formulation. Section \ref{section:TDKF} illustrates the proposed LTKF and TDKF state estimators for landmark tracking and relative localization, respectively. Experimental validation of the developed system is provided in section \ref{section:exp} and section \ref{section:conc} presents conclusions and future aspects of the developed work.

\begin{figure*}[ht]
\center
\includegraphics[width=\textwidth,height=0.55\textheight,scale=1]{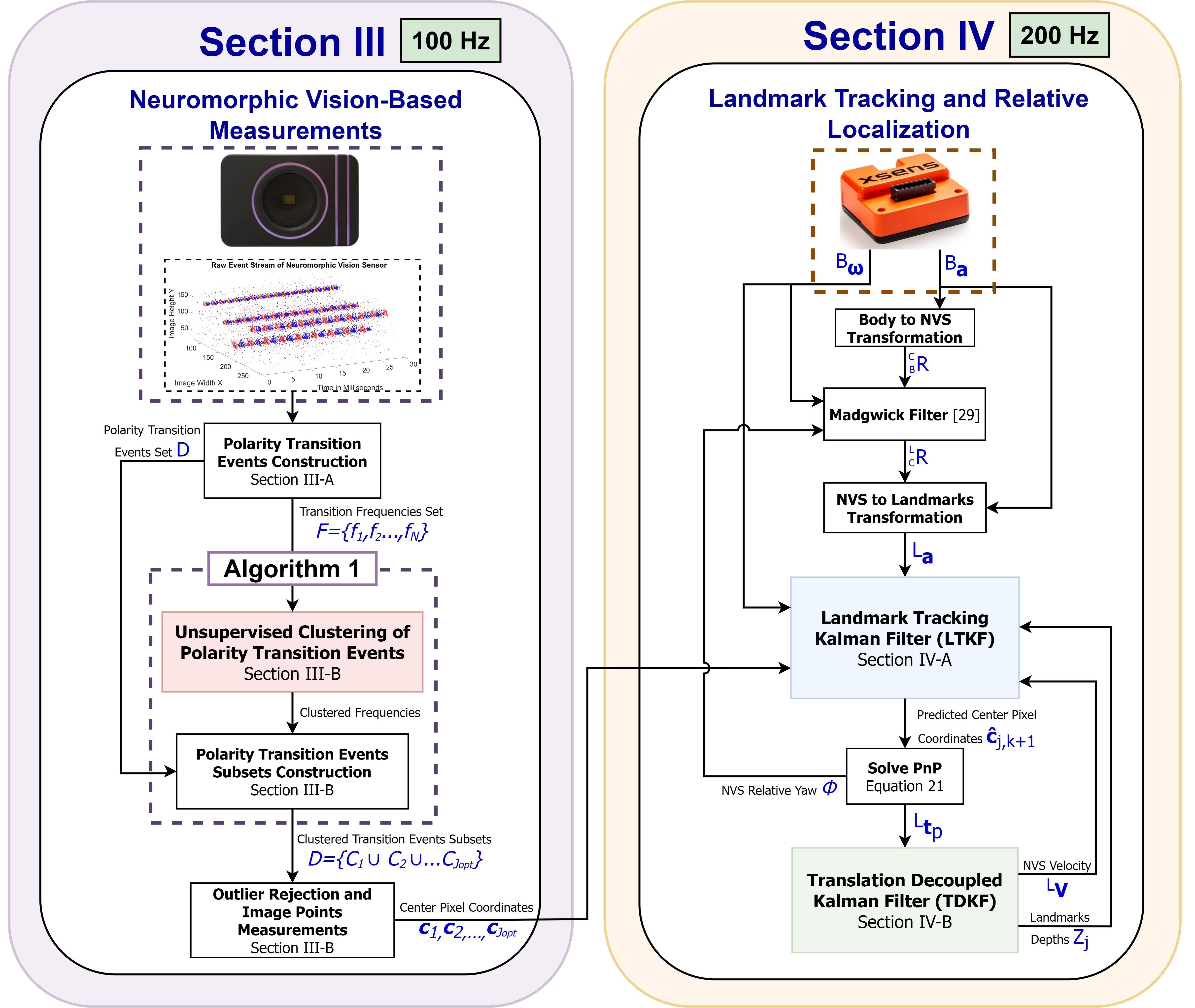}
\caption{Block diagram of the proposed relative localization system. The set of transition frequencies are clustered using GMMs and the landmarks pixel centers are extracted formulating the NVBMs at 100 Hz. The NVBMs are fed to the LTKF for tracking and relative localization is attained by the TDKF at 200 Hz.}
\label{fig:block_diagram}
\end{figure*}

\section{Preliminaries}
\label{section:overview}
\subsection{Notations and Reference Frames}
\begin{figure}[h]
\center
\includegraphics[scale=0.36]{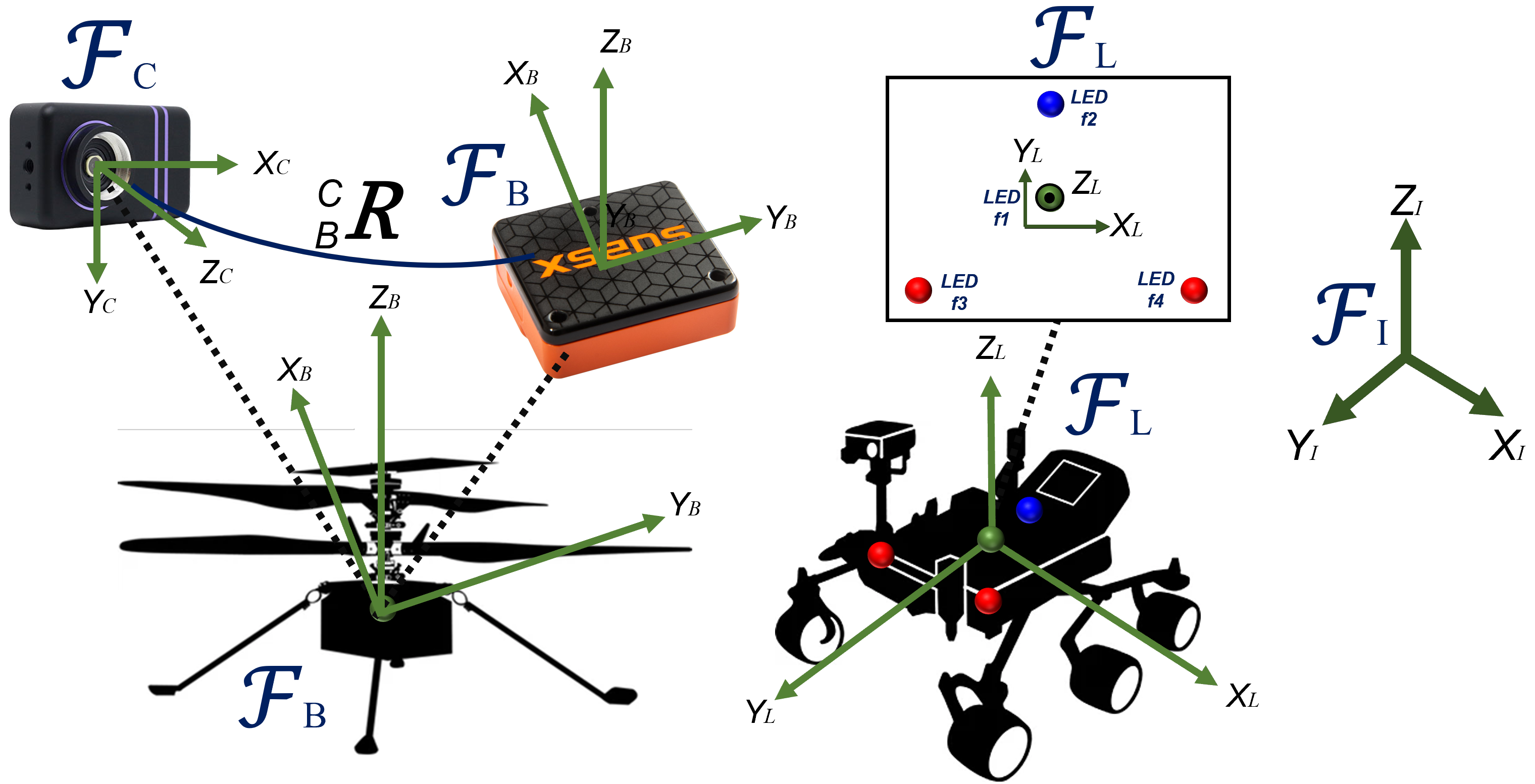}
\caption{The major frames of reference for the relative localization system.}
\label{fig:frames}
\end{figure}

The reference frames that are considered for developing our approach are shown in Fig. \ref{fig:frames}. An inertial frame \(\mathcal{F}_I\) is defined of basis \( \bm{[i_x, i_y, i_z]} \) with \(\bm{i_z}\) antiparallel to the gravity vector, and the IMU body fixed reference frame \(\mathcal{F}_B\). Also, the landmarks reference frame \(\mathcal{F}_L\) defining their relative positions with basis \( \bm{[l_x, l_y, l_z]} \), where \(\bm{l_z}\) is parallel to \(\bm{i_z}\). We also define the NVS frame of reference $\mathcal{F}_{C}$, with rotation matrix \(  {}^C_B\bm{R} = \bm{[b_x, b_y, b_z]} \in \text{SO(3)} \), giving the transformation from \(\mathcal{F}_B\) to \(\mathcal{F}_C\). A vector in a particular reference frame is denoted with a pre-superscript, e.g. \({}^I\bm{t}\) is the translation vector expressed in the inertial frame. In addition, components of vectors are denoted with subscripts (i.e. \({}^I\bm{t}=[{}^It_x {}^It_y {}^It_z]^T\)).

\subsection{System Overview} \label{subsection:system_overview}
A block diagram of the proposed system is shown in Fig. \ref{fig:block_diagram} relying on measurement streams from the NVS and IMU. The system comprises of two major subsystems for landmark identification and tracking, and relative localization, respectively. Transition frequencies from the raw event stream are constructed since they provide a better representation of the landmarks carrier waves. The set of transition frequencies are clustered using GMMs to extract the true frequencies of the flickering landmarks as identifiers due to the high levels of Gaussian noise accompanied with the transition frequency measurements. Then, polarity transition events subsets are formulated given the frequency-based cluster assignments where blob detection is utilized for outlier rejection and extracting the landmarks 2D center pixel coordinates on each all events of each cluster formulating our NVBMs. One key aspect of the proposed system is that the landmarks flickering frequencies were selected to be greater than 200 Hz to avoid correlation with low frequency background clutter. Accordingly, 100 Hz was selected as the frequency detection resolution to reliably detect all the landmarks frequencies as well as rejecting any frequency below 100 Hz. Nevertheless, we utilize the LTKF for tracking the landmarks based on their estimated pixel velocities running at the IMU higher update rate (200 Hz) providing robust tracking even when harsh maneuvers are performed.

It is important to note that to estimate pixel velocities, the Ingenuity linear and angular velocities along with the landmarks depths in the $\mathcal{F}_{C}$ need to be given as a priori. Therefore, we propose the TDKF running at also 200 Hz fusing relative pose estimates from a solution to the perspective-n-point (PnP) problem $^{L}\mathbf{t}_{p}$ and acceleration measurements $^{B}\mathbf{a}$ received from the IMU to estimate the Ingenuity linear velocity and its relative pose. We assume that the gyroscope bias is negligible and the angular velocity measurements are utilized directly in the LTKF. Surely, the IMU acceleration measurements need to be transformed to \(\mathcal{F}_L\), where the transformation is obtained by the Madgwick filter and yaw angle estimates are utilized from the solve PnP block to account for yaw angle drift. We apply the proposed system on relative localization between the Ingenuity and Perseverance Rover, where $\mathcal{F}_{B}$ also resembles the Ingenuity body-fixed reference frame and $\mathcal{F}_{L}$ is aligned with rover body frame of reference as shown in Fig. \ref{fig:frames}.

\section{Neurmorphic Vision-Based Measurements}
\label{section:LTKF}
\subsection{Flickering Landmarks Frequency Detections} \label{subsection:frequency_detections}

The NVS consists of an array of pixels each consisting of a photodiode, generating photocurrents stimulated by arriving photons, which get processed asyncronously and independently for every pixel. Every pixel circuit produces a binary output (i.e. polarity) corresponding to the log intensity changes. These binary outputs get spatial and temporal stamp producing an event
\begin{equation}
    e_k\doteq(\bm{x}_k,t_k,p_k)
\end{equation}
with \(\bm{x}\in W\times H\) being pixel location where \(W\in\{1,...,w\}\) and \(H\in\{1,...,h\}\) representing an image of \(h\) pixels in height and \(w\) pixels in width, \(t\) is the timestamp, and \(p\in\{-1,+1\}\) is the polarity of the event. Let a \emph{polarity transition event}, \(\Delta_k\), be defined as:
\begin{equation}
    \Delta_k\doteq  e_k-e_l : \bm{x}_k \equiv  \bm{x}_l \land t_k-t_l<\tau/2 \land p_k-p_l= 2
\end{equation}
where the difference operation \(e_k-e_l\) is defined as a substraction of the timestamp and polarity elements (i.e. \(\bm{x}_k\) is the same for \(\Delta_k\), \(e_k\) and \(e_l\)), and \(\tau\) represents the maximum period that is considered for conecutive events. Note that it is assumed that \(t_k>t_l\). The \emph{polarity transition frequency} is defined as \(f_k\doteq  (2\Delta_k)_2^{-1}\), where the notation \((.)_i\) is used to reference the \(i^{th}\) element in the tuple. The polarity transition events \(\Delta\) and frequencies \(f\) are consecutively added to the sets \(D\), \(\{\Delta_k\}\cup D\), and \(F\), \(\{f_k\}\cup F\), respectively. The sets \(D\) and \(F\) store all \(\Delta\) and \(f\) happened in the previous \(\tau\) seconds. Hence \(f_d=\tau^{-1}\) is called the \emph{frequency detection resolution} which we select, in this work, to be 100 Hz (i.e. \(\tau=10\) ms) as described in section \ref{subsection:system_overview}. As a consequence, polarity transition events with frequencies less than \(f_d\) cannot be detected.

The set $F$ is the input to Algorithm \ref{alg:gmm}, and the output is \(C_1\cup C_2\cup ... \cup C_{J_{opt}} = D\) clustered sets with \(j\in\{1,2,...,J_{opt}\}\). The optimal number of GMM clusters \(J_{opt}\) corresponding to the number of observed flickering landmarks is selected by minimizing the Bayesian information criterion (BIC) using the method detailed in the following Section and summarized in Algorithm \ref{alg:gmm}.

\subsection{Unsupervised clustering of polarity transition events} \label{subsection:nvbm}
The clustering of polarity transition events represented in the set \(D\) is performed using the set of transition frequencies in \(F\), those we ignore any spatial information about the polarity transition events during clustering.

\begin{figure}[t]
\center
\includegraphics[scale=0.26]{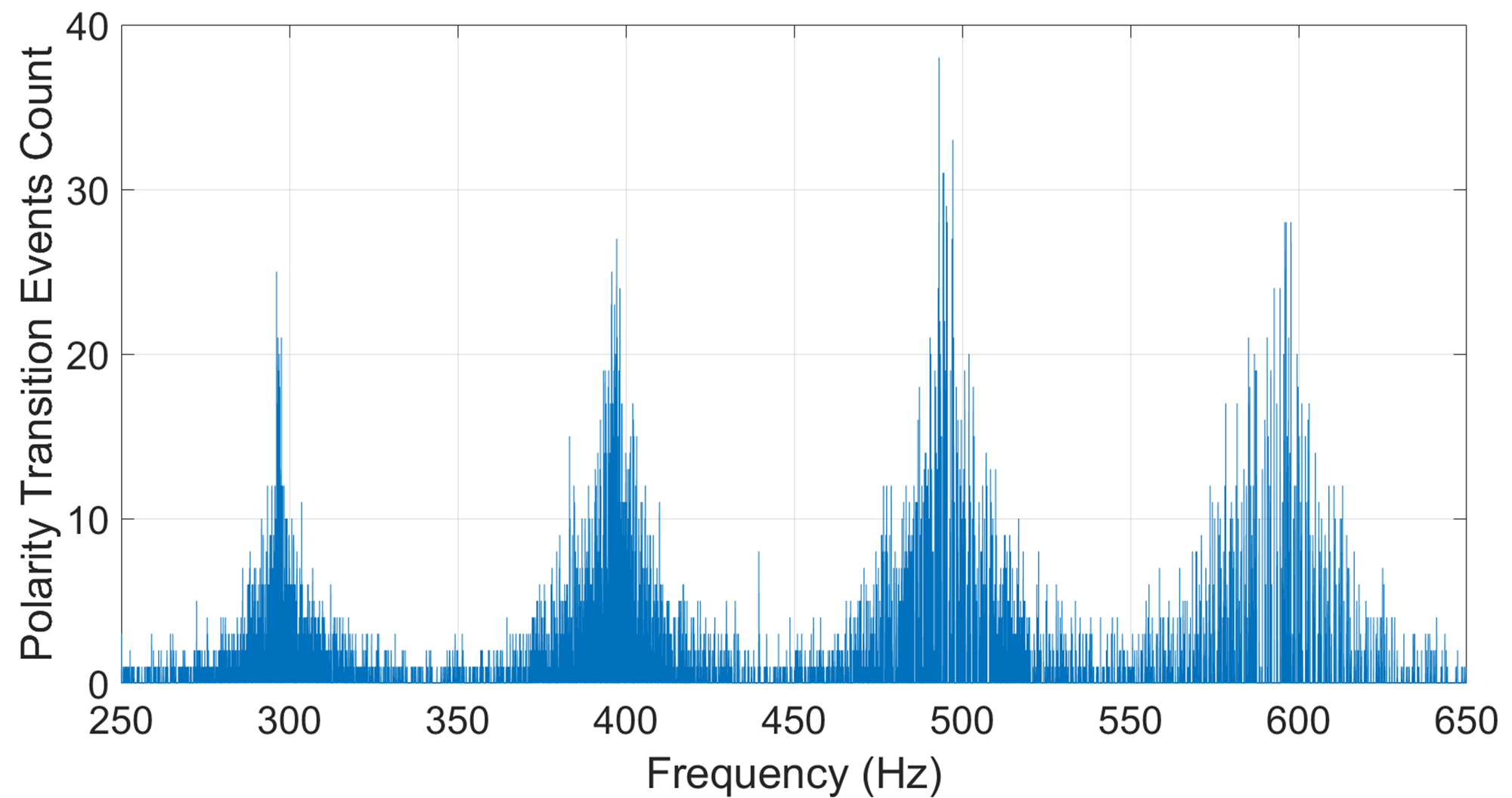}
\caption{A plot of the transition frequencies histogram following a Gaussian mixture model.}
\label{fig:noisy_f}
\end{figure}

Algorithm \ref{alg:gmm} specifies the procedure of clustering the transition frequencies in a set $F$ for landmark identification. If we plot a histogram of $F$ comprising of 172 transition frequency elements for four observed flickering landmarks with frequencies 300, 400, 500, and 600 Hz, shown in Fig. \ref{fig:noisy_f}, it is obvious that the probability distribution of the transition frequencies $p(f_{k})$ is dominated by a mixture of Gaussians, see Fig. \ref{fig:noisy_f}. Hence, $p(f_{k})$ is parameterized by the mixture components' mean $\mu_{j}$, covariance $\Sigma_{j}$, mixture proportion $\pi_{j}$, and is modelled by
\begin{equation}
    p(f_k) = \sum_{j=1}^{J} \mathcal{N}(f_k | \mu_{j},\Sigma_{j})\pi_j
\end{equation}
where $J$ represents the current number of clusters corresponding to the number of observed landmarks by the NVS.

The optimal values of $\mu_j$, $\pi_j$, and $\Sigma_{j}$ are the ones minimizing the $p(f_{k})$ expected likelihood $L$ \cite{ml_bishop} defined as

\begin{equation}
    L = \sum_{n}^{N}\sum_{j}z_{n}^{j}\log\pi_{j} - \sum_{n}^{N}\sum_{j}^{J}z_{n}^{j}\frac{1}{2\Sigma}(f_{n}-\mu_{j})^{2}
\end{equation}
\noindent using the expectation maximization (EM) algorithm, where $N$ represents the number of elements in $F$.

The EM algorithm comprises of two iterative computations that are repeated until convergence: (1) the expectation (E) step and (2) maximization (M) step. The E step evaluates the responsibilities $\beta(z_{nj})$ defined as

\begin{equation}
    \beta(z_{nj}) = \frac{\pi_{j}\mathcal{N}(f_{k}|\mu_{j},\Sigma_{j})}{\sum_{j}\pi_{j}\mathcal{N}(f_{k}|\mu_{j},\Sigma_{j})}.
    \label{equation:beta}
\end{equation}

\noindent On the other hand, the M step revaluates $\mu_j$, $\pi_j$, and $\Sigma_{j}$ by

\begin{equation}
    \mu_{j} = \frac{\sum_{n}\beta(z_{nj})f_{k}}{\sum_{n}\beta(z_{nj})},
    \label{equation:mu}
\end{equation}

\begin{equation}
    \Sigma_{j} = \frac{1}{\sum_{n}\beta(z_{nj})}\sum_{n}\beta(z_{nj})(f_{n} - \mu_{j})(f_{n} - \mu_{j})^{T},
    \label{equation:sigma}
\end{equation}

\begin{equation}
    \pi_{j} = \frac{\sum_{n}\beta(z_{nj})}{N}.
    \label{equation:pi}
\end{equation}

It is important to note that the EM algorithm requires $J$ to be known as a priori which is not practical in the targeted application given the limited FOV of the NVS. Therefore, $J_{opt}$ is selected based on minimizing the BIC $\gamma$ expressed as \cite{ml_bishop}

\begin{equation}
    \gamma = N \cdot \ln(\hat{\Sigma_{e}^{j}}) + \alpha \cdot \ln(N),
    \label{equation:bic}
\end{equation}

\noindent where $\alpha$ is 4 corresponding to the number of parameters to be estimated, $\mu_{j}$, $\Sigma_{j}$, $\pi_{j}$, and $J$, and $\hat{\Sigma_{e}^{j}}$ is the error covariance defined as 

\begin{equation}
    \hat{\Sigma_{e}^{j}} = \frac{1}{N}\sum_{j}\sum_{n}(f_{n} - \mu_{j}).
\end{equation}

\noindent The $\gamma$ is evaluated on cluster hypotheses $i$ ranging from 1 to $J_{max}$, which we select as 10 to maintain the algorithm's computational efficiency.

Once the polarity transition events have been assigned to clusters, every element \(f_n\) is associated with the corresponding element \(\Delta_k\). With such association, it is possible to filter out events that are not in the spatial neighborhood of the majority of the events in the cluster, and we achieve that by performing blob detection followed by masking the largest blob resulting in a binary image \(M_j\) \cite{blob}, where \(j\in\{1,...,J_{opt}\}\). The moment of each image \(M_j\) is used to find the center pixel coordinates, \(\bm{c}_j=[u \; v]^{T}\), corresponding to each flickering landmark.

\begin{algorithm}[t]
\caption{Unsupervised Clustering of Polarity Transition Events}
\begin{algorithmic}[1]
\Inputs{Set $F = \{ f_{1},f_{2},...,f_{N} \}$} \\
\textbf{Outputs}: {Clustered sets: \(D = C_1\cup C_2\cup ... \cup C_{J_{opt}}\)}
\Initialize{\strut $p(f_{k}) = \sum_{j=1}^{J} \mathcal{N}(f_{k} | \mu_{j},\Sigma_{j})\pi_j$}
\For{Each set $F$ received}
\For{$i$ = $1$ to $J_{max}$}
\State Initialize $\mu_{j}$, $\Sigma_{j}$, $\mu_{j}$
\While{Not converged}
\State \textbf{E step}: evaluate: $\beta(Z_{n})$ ; Eq. \ref{equation:beta}
\State \textbf{M step} evaluate: $\mu_{j}$, $\Sigma_{j}$, $\pi_{j}$ ; Eqs. \ref{equation:mu}, \ref{equation:sigma}, \ref{equation:pi}
\State Check for convergence
\EndWhile
\EndFor
\State Find $J_{opt}$ for minimum $\gamma$
\EndFor
\end{algorithmic}
\label{alg:gmm}
\end{algorithm}

\section{Landmark Tracking and Relative Localization}
\label{section:TDKF}
\subsection{Landmark Tracking Kalman Filter} \label{subsection:ltkf}
The center pixel coordinates \(\bm{c}_j\) corresponding to flickering landmarks in the image plane got estimated at the rate \(f_d\), but it is possible to increase the rate of estimation to \(f_i\) corresponding to the NVS inertial states estimation rate. The center pixel velocities \(\dot{\bm{c}}_j\) can be estimated through the knowledge of the NVS inertial states and the relative location of the flickering landmarks. This requires the assumption that the landmarks are quasi static. The number of LTKFs used is \(J_{opt}\). When a flickering landmark goes out of the FOV, its corresponding LTKF gets removed, and it get initialized again only once the corresponding flickering landmarks gets in the FOV again.

Thus, to design the LTKF, we use the center pixel velocities in the LTKF prediction model to estimate the landmarks image points, and we use the \(\bm{c}_j\) measurements in the correction step. The relative velocities of the landmarks 3D points ${}^{C}\mathbf{\Dot{X}}_{j} = [ \dot{x} \text{ } \dot{y} \text{ } \dot{z} ]^{T}$ are directly related to the NVS linear velocity ${}^{L}\mathbf{v} = [ v_{x} \text{ } v_{y} \text{ } v_{z} ]^{T}$, angular velocity ${}^{L}\boldsymbol{\omega} = [ \omega_{x} \text{ } \omega_{y} \text{ } \omega_{z} ]^{T}$, and the relative depth of each landmark $Z_{j}$ defined in $\mathcal{F}_{L}$ by \cite{handbook}

\begin{equation}
    {}^{C}\mathbf{\dot{X}}_{j} = - {}^{L}\mathbf{v} - {}^{L}\boldsymbol{\omega} \times {}^{C}\mathbf{X}_{j},
\end{equation}
\noindent which expands to
\begin{align}
    \Dot{x} &= -v_{x} -\omega_{y}Z_{j} +\omega_{z}Y_{j} \label{equation:velsx}, \\
    \Dot{y} &= -v_{y} -\omega_{z}X_{j} +\omega_{x}Z_{j} \label{equation:velsy}, \\
    \Dot{z} &= -v_{z} -\omega_{x}Y_{j} +\omega_{y}X_{j}, \label{equation:velsz}
\end{align}

\noindent where $X_{j}$, $Y_{j}$, and $Z_{j}$ are the landmarks relative positions defined in $\mathcal{F}_{C}$.

Accordingly, the landmarks pixel velocities $\dot{\mathbf{c}}_{j} = [ \dot{u} \text{ } \dot{v} ]^{T}$ are a function of their center pixel coordinates $\mathbf{c}_{j}$, the relative depth of the landmarks $Z_{j}$, and the camera intrinsic matrix $K$ as \cite{handbook,arxiv_servoing}

\begin{equation}
    \dot{\mathbf{c}}_{j} = \mathbf{L}_{s,j}(u,v,Z_{j},K)
    \begin{bmatrix}
    {}^{L}\mathbf{v} \\
    {}^{L}\boldsymbol{\omega}
    \end{bmatrix},
    \label{equation:pixel_vels}
\end{equation}

\noindent where $\mathbf{L_{s,j}}$ $\in$ $\mathbb{R}^{2\times6}$ is the interaction matrix for each landmark. Accordingly, taking the time derivative of $\Dot{u}_{i}$, the interaction matrix for each flickering landmark $\mathbf{L_{s}}$ is given by \cite{arxiv_servoing}

\begin{equation*}
    \mathbf{L_{s}} = \begin{pmatrix}
    -\frac{f_{u}}{Z_{j}} & 0 & \frac{u}{Z_{j}} & \frac{uv}{f_{v}} & -f_{u}-\frac{u^{2}}{f_{u}} & \frac{f_{u}}{f_{v}}v \\
    0 & -\frac{f_{v}}{Z_{j}} & \frac{v}{Z_{j}} & f_{v} + \frac{v^{2}}{f_{v}} & -\frac{uv}{f_{u}} & -\frac{f_{u}}{f_{v}}u
    \end{pmatrix}.
    \label{equation:interaction}
\end{equation*}

Through the nonlinear Eq. \ref{equation:pixel_vels} it is possible to fuse the measured \(\bm{c}_j\) with the inertial states of the NVS. However, due to lower update rate of \(\bm{c}_j\), pixelation errors, and noises in the polarity transition events, such fusion is neglected and \(\bm{c}_j\) is decoupled from NVS states in the LTKF. This would also allow us to have a computationally efficient filtration scheme by using a much smaller state vector. Accordingly, we define the prediction model for the LTKF as

\begin{align}
    \Tilde{\mathbf{c}}_{j,k+1} &= \begin{bmatrix}
     \Tilde{u}_{k+1} \\
     \Tilde{v}_{k+1}
     \end{bmatrix} \\ &= \begin{bmatrix}
    1 & 0 \\
    0 & 1 \\
    \end{bmatrix} \Tilde{\mathbf{c}}_{j,k} + \begin{bmatrix}
    \Delta t \\
    \Delta t \\
    \end{bmatrix} \mathbf{u}_{j,k} + \mathcal{N}(0,Q_{p}), \\
    & \text{where } j \in \{1,2,...,J_{opt}\}, \nonumber
\end{align}

\noindent where $\Tilde{\mathbf{c}}_{j,k}$ is the predicted state vector comprising the landmarks image points, $\mathbf{u}_{j,k}$ is the control input vector of the estimated pixel velocities $\dot{\mathbf{c}}_{j}$, and $Q_{p}$ is the covariance for an additive zero-mean Gaussian process noise. As customary for these types of filters, an update step is required for each of the LTKFs  in $j \in \{1,2,...,J_{opt}\}$, where their measurement models are defined with the measurement vectors $\mathbf{z}_{j,k} = \mathbf{c}_{j}$ and the observation matrices $H_{j,k} = \mathcal{I}_{2 \times 2}$. It is important to highlight that ${}^{L}\boldsymbol{\omega}$ is obtained from the IMU gyroscope and ${}^{L}\mathbf{v}$ is estimated from the TDKF, which is discussed in the following section.

\subsection{Translation Decoupled Kalman Filter} \label{subsection:tdkf}
Given the tracked landmarks image points $\Tilde{\mathbf{c}}_{j}$, it is possible to estimate the relative position of the NVS with respect to the landmarks by means of solutions to the PnP problem. Nevertheless, this does not provide the sufficient relative localization accuracy and the PnP positioning estimates need to be differentiated to obtain ${}^L\mathbf{v}$ which tends to be noisy, deteriorating the tracking performance of the LTKF. Therefore, the TDKF utilizes the IMU accelerometer measurements $\mathbf{}^{B}\mathbf{a} = [a_{x} \text{ } a_{y} \text{ } a_{z}]$ in its prediction model and the PnP positioning estimates as measurement corrections for robustly estimating the ${}^L\mathbf{v}$ as well as the translation vector ${}^{L}\mathbf{t}$ pointing from $\mathcal{F}_{C}$ to $\mathcal{F}_{L}$.

It is important that the acceleration measurements are defined in $\mathcal{F}_{L}$ to be utilized in the TDKF. First, $\mathbf{}^{B}\mathbf{a}$ is rotated to $\mathcal{F}_{C}$ with \({}^C_B\bm{R}\). The translational offset between frames $\mathcal{F}_{C}$ and $\mathcal{F}_{B}$ is neglected since the acceleration measurements are not affected by the NVS and IMU positions given that they are on the same rigid body. Then, $\mathbf{}^{C}\mathbf{a}$ is transformed to $\mathcal{F}_{L}$ using the Madgwick orientation filter \cite{madgwick}. Nevertheless, it is substantial to provide a reliable source for the relative yaw $\phi$ of $\mathcal{F}_{C}$ with respect to $\mathcal{F}_{L}$ to account for yaw angle drift. Hence, $\phi$ is estimated by solving the PnP problem replacing the magnetometer in Madgwick's filter \cite{madgwick}, demonstrated in Fig. \ref{fig:block_diagram}.

After obtaining ${}^{B}\mathbf{a}$ in $\mathcal{F}_{L}$, the TDKF prediction model is defined as follows

\begin{align}
    \Tilde{\mathbf{x}}_{k+1} &= \begin{bmatrix}
    {}^{L}\Tilde{\mathbf{t}}_{k} \\
    {}^{L}\Tilde{\mathbf{v}}_{k} \\
    \Tilde{\mathbf{a}}_{b,k}
    \end{bmatrix} \nonumber \\
    &= {\begin{bmatrix}
    1 & \Delta t & -\Delta t^{2} \\
    0 & 1  & -\Delta t \\
    0 & 0 & 1
    \end{bmatrix}}\Tilde{\mathbf{x}}_{k} + {\begin{bmatrix}
    \Delta t^{2} \\
    \Delta t \\
    0
    \end{bmatrix}}u_{k}  + \mathcal{N}(0,\sigma_{p}), \label{equation:predicted_state}
\end{align}

\noindent where $\Tilde{\mathbf{x}}_{k+1}$ is the TDKF predicted state vector comrpising of the predicted NVS position ${}^{L}\Tilde{\mathbf{t}}_{k}$ and velocity ${}^{L}\Tilde{\mathbf{v}}_{k}$ along each axis, $a_{b,k}$ is an estimate for acceleration bias to adjust for gravity, and $Q_{p}$ is the variance for an additive zero-mean Gaussian process noise.

The prediction model of the TDKF provides reliable relative pose estimates but measurement corrections are substantial to avoid drifts due to the integration of acceleration bias. Thus, solutions to the PnP problem providing are used to correct the predicted TDKF states.

Solutions to the PnP problem estimate the translation ${}^{L}\mathbf{t}_{p}$ and rotation ${}_{C}^{L}R_{p}$ of $\mathcal{F}_{NVS}$ with respect to $\mathcal{F}_{L}$ given ${}^{L}\mathbf{X}_{j}$ projected as $\mathbf{c}_{j}$ with the NVS perspective projection model as

\begin{equation*}
        s\begin{bmatrix}
        \mathbf{c}_{j} \\
        1
    \end{bmatrix} = \underbrace{
    \begin{bmatrix}
f{x} & \gamma & u_{0}\\
0 & f_{y} & v_{0}\\
0 & 0 & 1
\end{bmatrix}
    }_{K} \underbrace{
    \begin{bmatrix}
r_{11} & r_{12} & r_{13} & t_{1}\\
r_{21} & r_{22} & r_{23} & t_{2}\\
r_{31} & r_{32} & r_{33} & t_{3}
\end{bmatrix}}_{{}_{C}^{L}R_{p}|{}^{L}\mathbf{t}_{p}} \begin{bmatrix}
{}^{L}\mathbf{X}_{j} \\
1
\end{bmatrix},
\label{equation:projection}
\end{equation*}

\noindent where $s$ is a scaling factor. Obviously, no algebraic solution exists for ${}_{C}^{L}R_{p}$ and ${}^{L}\mathbf{t}_{p}$, since for each projected point the number of unknown parameters exceeds the known terms. Hence, the Gauss Newton minimization scheme (EPnP-GN) \cite{epnp} is utilized to estimate both parameters obtained by iteratively minimizing the reprojection error, where the states are paramaterized with an exponential map and the perspective projection model transforms to

\begin{equation}
        s\begin{bmatrix}
        \mathbf{c}_{j} \\
        1
    \end{bmatrix} = K \exp({\zeta})
    \begin{bmatrix}
    {}^{L}\mathbf{X}_{j} \\
    1
\end{bmatrix}.
\label{equation:cam_proj}
\end{equation}

\noindent From Eq. \ref{equation:cam_proj}, a least-squares problem is formulated to minimize the reprojection error as a function of $\zeta$:

\begin{equation}
    \zeta = \argmin{}_{\zeta}\frac{1}{2}\sum_{i}||u_{i} -\frac{1}{s}K\exp{(\zeta)} X_{w}||^{2}
    \label{equation:epnp_optim}
\end{equation}

\noindent After obtaining ${}^{L}\mathbf{t}_{p}$ and ${}_{C}^{L}R_{p}$, the TDKF measurement model is formulated with the measurement vector $\mathbf{z}={}^{L}\mathbf{t}_{p}$ and observation matrix $H_{k}=[\mathcal{I}_{3 \times 3} \text{ } 0_{3 \times 6}]$.

\section{Experiments}
\label{section:exp}
\subsection{Experimental Setup}

\begin{figure*}[ht]
\center
\includegraphics[width=\textwidth,height=\textheight,keepaspectratio,scale=1]{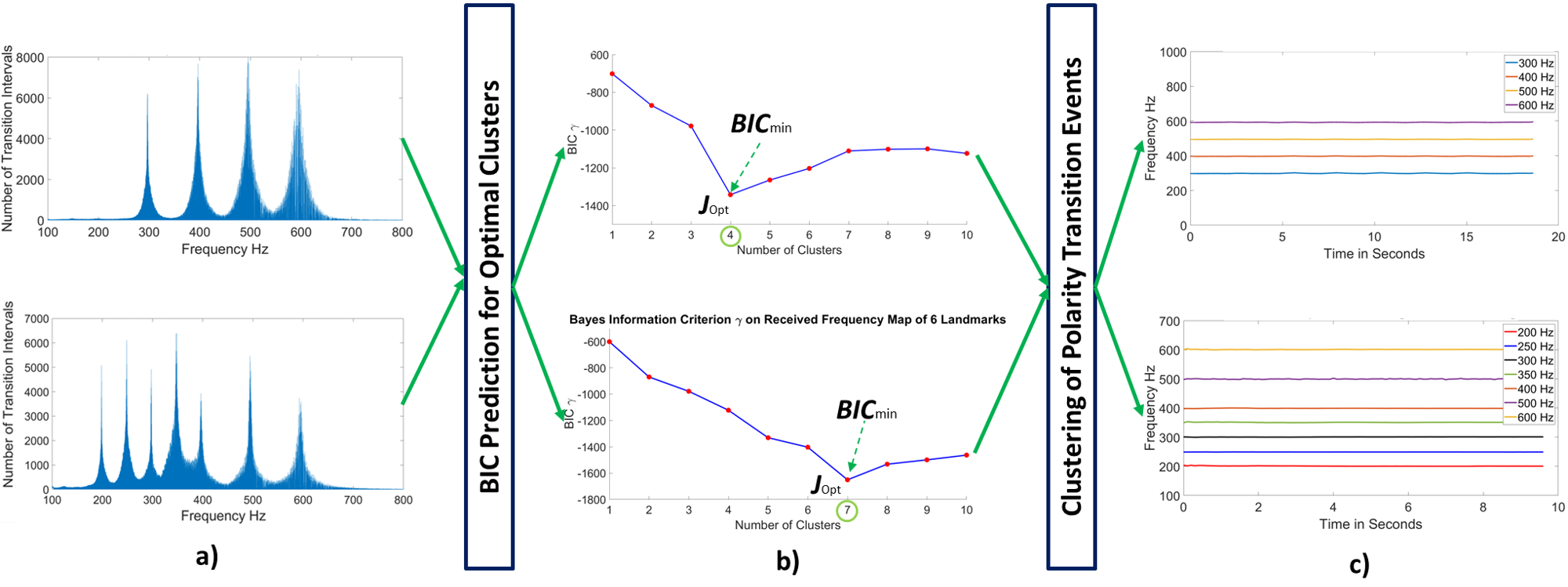}
\caption{a) The detected noisy polarity transition frequencies. b) The polarity transition events are clustered using GMMs and the BIC criterion extracts the optimal number of clusters corresponding to the observed number of landmarks. c) The clustered transition frequencies represent the true value of the landmarks flickering frequencies.}
\label{fig:f_filter}
\end{figure*}

\begin{figure}[t]
\center
\includegraphics[scale=0.7]{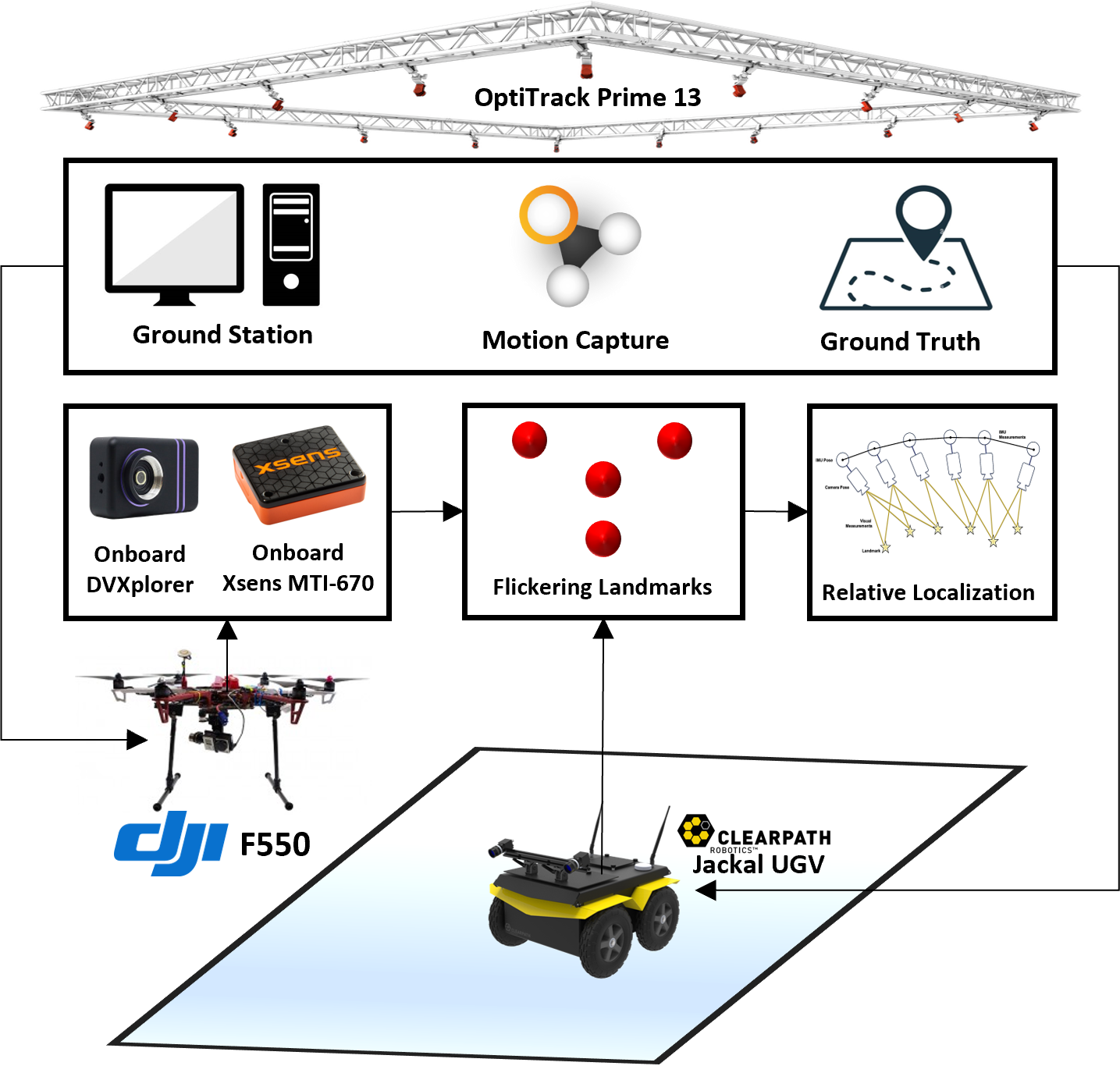}
\caption{Experimental protocol followed comprising of DJI F550 with onboard NVS and IMU and Jackal with flickering landmarks.}
\label{fig:exp_setup}
\end{figure}

\begin{figure}[h]
\center
\includegraphics[scale=0.4]{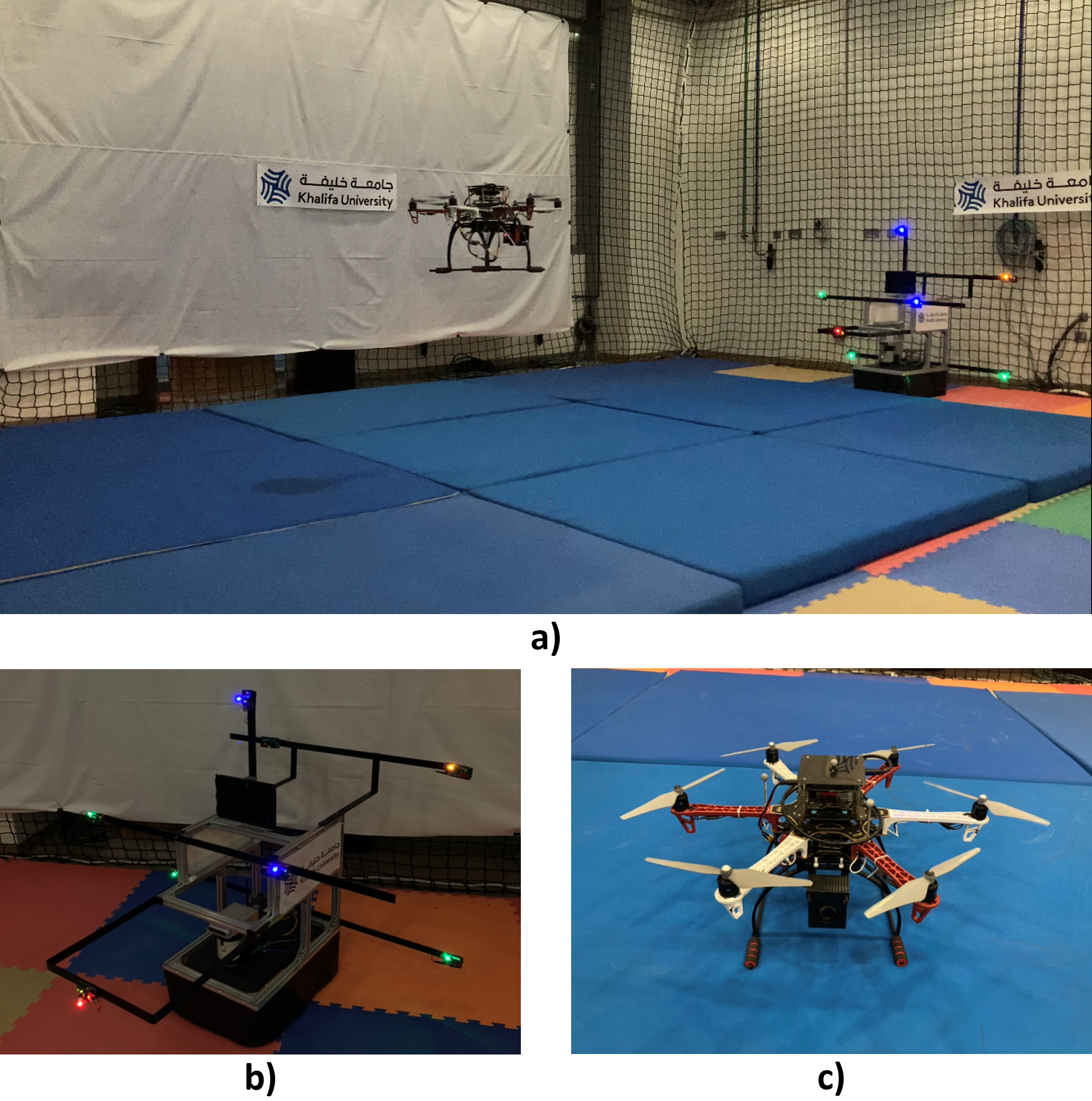}
\caption{a) Flying UAV observing flickering landmarks mounted on the UGV, b) Jackal with seven landmarks, and c) DJI F550 with DVXplorer and Xsens MTI-670 IMU.}
\label{fig:drone_ugv}
\end{figure}

We applied our approach on a space application to perform relative localization between the Ingenuity and the Perseverance Rover. The Ingenuity shall be represented by an unmanned aerial vehicle (UAV) with onboard NVS and IMU, and the rover by an unmanned ground vehicle (UGV) with the flickering landmarks; the experimental architecture is demonstrated in Fig, \ref{fig:exp_setup}. Accordingly, DJI F550 and Jackal were utilized as a UAV and a UGV, respectively, see Fig. \ref{fig:drone_ugv}. The experiments were performed on two scenarios: (1) UAV performing a take off and hover and (2) UAV following a square trajectory while observing the landmarks on the UGV. The experiments were carried out within a seven meter range due to the limited indoor space envelope. Videos of the experiments are available through the following link: \url{https://youtu.be/N3Ypu95c3_4}.

It is important to highlight that the UGV was assumed static since the Perseverance Rover speed (0.043 m/s) is relatively slow compared to the Ingenuity (10 m/s) \cite{copter_demonstrator}. This also satisfies the requirement on quasi static features given in section \ref{section:LTKF}. This assumption holds given the update rate of the LTKF measurement corrections (100 Hz), in which rising positioning error due to the assumption would be approximately 0.0004 m. Nevertheless, this can be easily surpassed with a constant velocity model adopted for the UGV. On the other hand, an Arduino Uno R3 ATmega16U2 was used to deliver unique frequency PWM signals to each landmark. Seven LedTech 228-5073 luminaries, of luminous intensity of 1300 mcd, were used as features flickering at 200, 250, 300, 350, 400, 500, and 600 Hz. The landmarks were placed at 1 meter apart abiding by the Perseverance Rover dimensions (3 m $\times$ 3 m $\times$ 3 m). This is also substantial for robust relative localization and avoiding near planar constellations which introduce positioning flip ambiguities. The DVXplorer was utilized as the NVS and Xsens MTI-670 as the IMU, both interfaced with robot operating system (ROS) using a USB 3.0 terminal on the UAV onboard processor. Two important aspects that were taken into consideration during the experiments are the time and space asynchrony between the received measurement streams from the NVS and the IMU. The asynchronous measurements were synchronized using soft time synchronization by ROS \cite{msg_filters} providing timestamped measurements from both sensors. The time offset between the two sensor clocks needed to be specified as a priori and was obtained through a step test with the times in which both sensors indicate that the UAV started flying have been compared. The time offset has been estimated to be 3.2 ms. On the other hand, the rotation between the NVS and IMU frames is substantial for the robustness of the system as discussed in section \ref{subsection:tdkf}. This was obtained by converting NVS events to grayscale images using \cite{e2vid} and utilizing the approach in \cite{kalibr} to obtain the inertial measurements in the NVS reference frame. Finally, Optitrack Prime 13 was utilized to provide ground truth poses for both platforms.

\begin{figure*}[ht]
\center
\includegraphics[width=\textwidth,height=\textheight,keepaspectratio,scale=1]{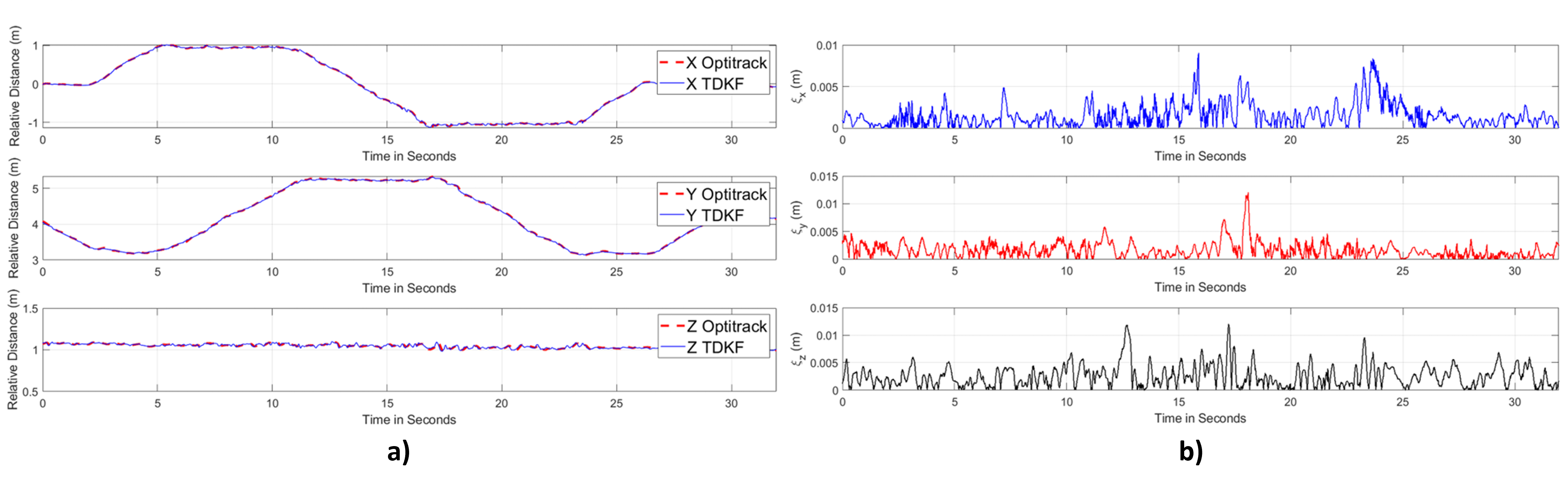}
\caption{a) UAV position relative to UGV estimated by TDKF against optitrack, where a maximum absolute positioning error $\xi_{x,y,z}^{max}$ of 0.0137 m and b) a maximum absolute relative orientation error $\xi_{\phi,\theta,\psi}^{max}$ of 2.16 degrees was achieved.}
\label{fig:motion}
\end{figure*}

\subsection{Landmark Identification Results}
Accurate frequency-based identification is critical to accurately formulate the NVBMs. The frequency detection error can be up to 33.97 Hz when relying on raw frequency measurements for landmark identification. This can lead the PnP solutions to diverge as the landmarks' correspondences are not matched correctly. On the other hand, we show that the unsupervised clustering of polarity transition events algorithm extracts the landmarks' true flickering frequencies and learns the optimal number of clusters for the polarity transition events. This was validated on the previously mentioned experimental setups and was also tested in two more scenarios where a random hand-motion was induced on (1) four and (2) seven observed landmarks, illustrated in Fig. \ref{fig:f_filter}. When relying on our unsupervised clustering approach, the maximum frequency detection error dropped to 3.21 Hz. This is surely robust since the landmarks correspondences are matched correctly and the PnP problem solution is optimized. In addition, the minimum $\gamma$ proved to be sufficient for finding $J_{opt}$ corresponding to the number of observed landmarks.

\subsection{Landmark Tracking Results}
\begin{figure}[h]
\center
\includegraphics[scale=0.26]{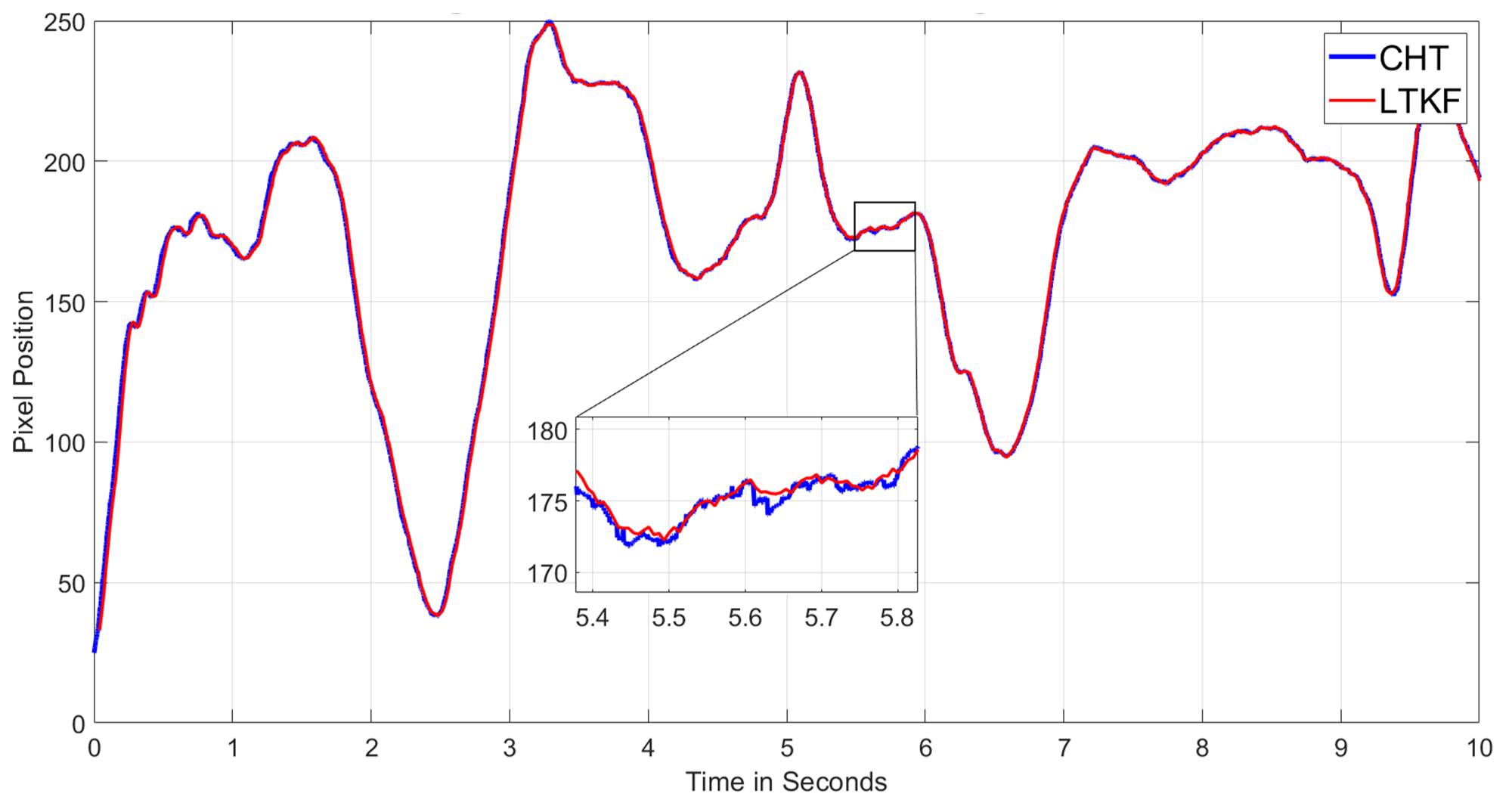}
\caption{Tracking plots for the CHT and LTKF on a 600 Hz landmark along X.}
\label{fig:tracking_plot}
\end{figure}

The LTKF provides a smoother and more robust landmark tracking especially in agile maneuvers compared to conventional imaging techniques relying on tracking by instance detection. We compare the LTKF to one of the most prominent feature extraction techniques for vision-based measurements, circle hough transform (CHT) \cite{cht}, running at 100 Hz during take-off in the hovering scenario, see Fig. \ref{fig:tracking_plot}. We rely on linear interpolation for their comparison due to their heterogeneous update rates. Note that the huge fluctuations are due to the non-optimal tuning of the UAV which caused it to vibrate.

The average difference between both trackers' outputs for a 600 Hz landmark along the Y axis in the image plane is 0.78 pixels which provides the required performance for robust relative localization between both platforms. It is still important to highlight that the LTKF is delayed compared to the CHT due to its inherent dynamics but does not exceed 0.0072 $ms$, which is negligible compared to its 5 $ms$ sampling time.

\subsection{Relative Localization Results}

\begin{figure}[h]
\center
\includegraphics[scale=0.26]{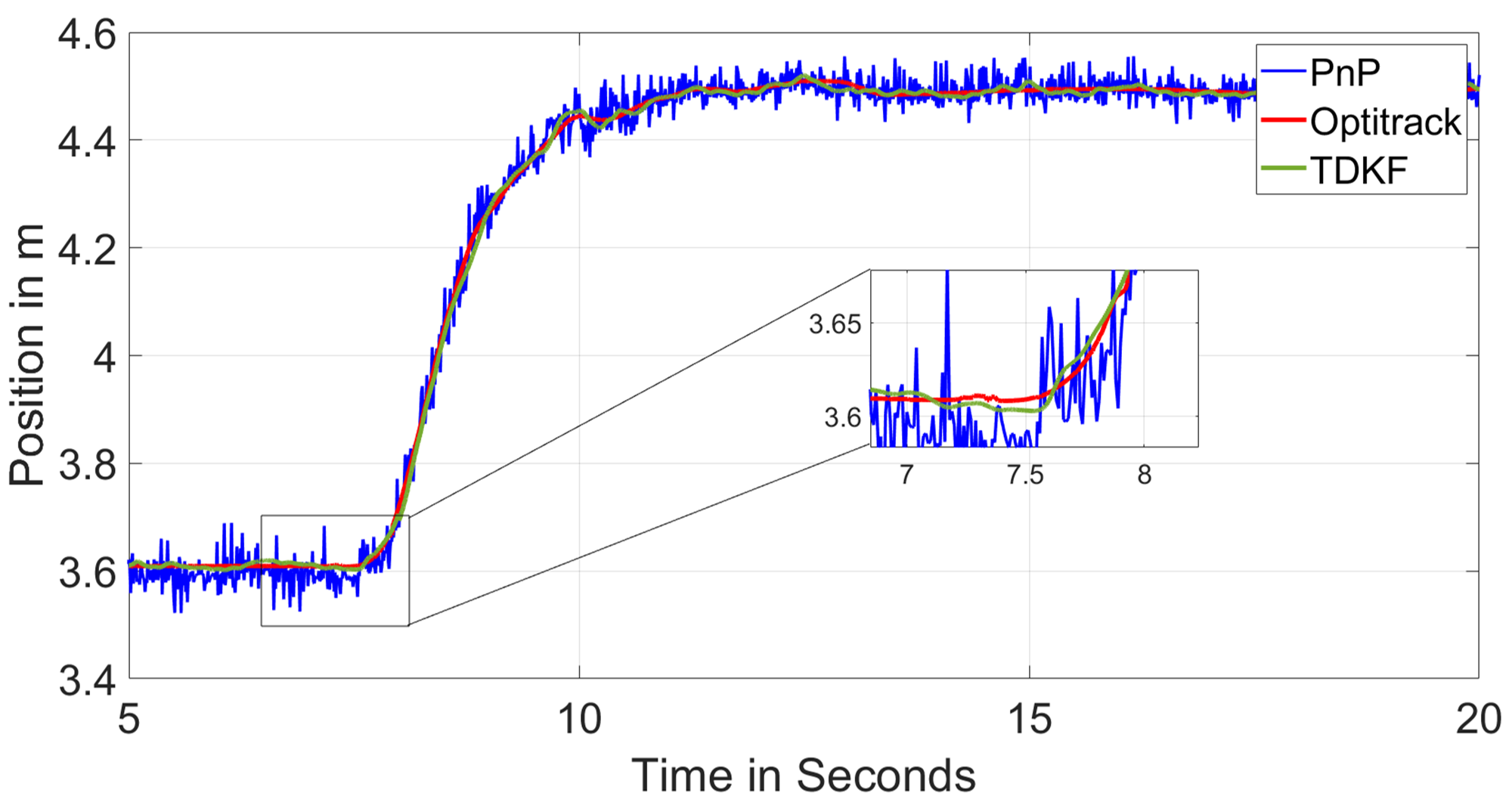}
\caption{Predicted relative position of PnP, TDKF, and optitrack demonstrating the noise present in CHT-PnP estimates and are filtered with the TDKF.}
\label{fig:pnp_nrkf}
\end{figure}

We validate the accuracy of our TDKF against the ground truth from the optitrack motion capture system for the UAV's relative position to the UGV. We first show that relative positioning estimates of the CHT-PnP suffers from large positioning errors compared to the Optitrack and the LTKF-TDKF state estimators, illustrated in Fig. \ref{fig:pnp_nrkf}.

\begin{figure}[h]
\center
\includegraphics[scale=0.25]{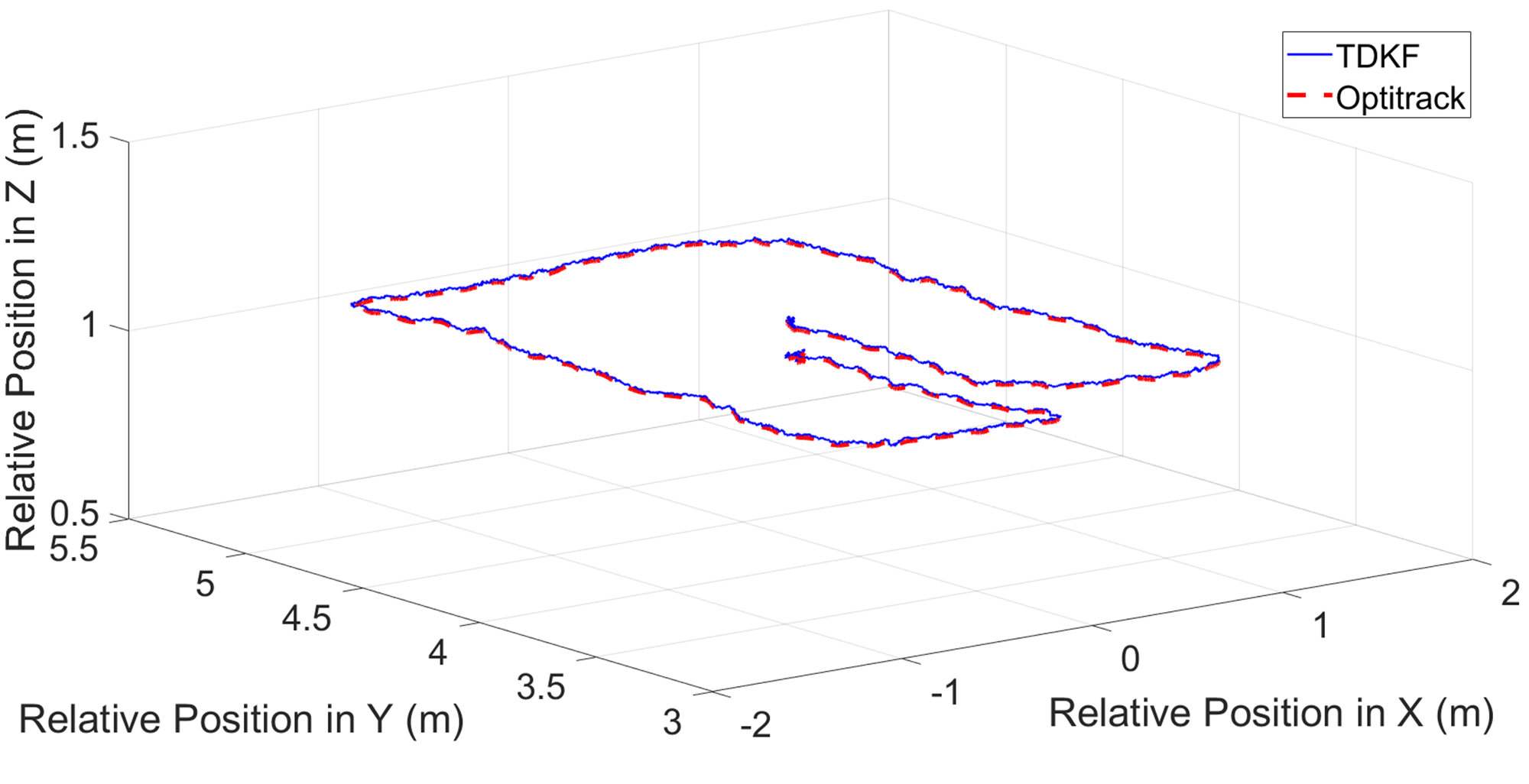}
\caption{Estimated UAV trajectory by the TDKF compared against optitrack.}
\label{fig:trajectory}
\end{figure}

By only solving the PnP problem for relative localization, the positioning error can rise up to 0.09 m. This is due to the noise present in the CHT tracking performance and the relatively low resolution of the NVS. On the other hand, the LTKF-TDKF filters the noise providing more robust positioning estimates. Fig. \ref{fig:motion} shows the TDKF estimated relative position compared to Optitrack for the scenario in which the UAV was following a square trajectory and Fig. \ref{fig:trajectory} shows the trajectory followed relative to the UGV. Accordingly, a maximum relative positioning error $\xi_{x,y,z}^{max}$ of 0.0137 m and a mean relative positioning error $\bar{\xi}_{x,y,z}$ of 0.0052 m within 7 meter range were achieved with the TDKF. It is noticeable that the positioning error tends to increase at time instances where the UAV starts to move along an axis. This rises from the sudden induced motion (i.e. at $t$ = 17 s in Y) but the TDKF converges quickly, yet the error remains within the acceptable range. On the other hand, the Madgwick filter fused with the yaw measurements obtained from the PnP solution was utilized for estimating the UAV orientation, achieving maximum orientation error $\xi_{\phi,\theta,\psi}^{max}$ of 2.16 degrees and a mean orientation error $\bar{\xi}_{\phi,\theta,\psi}$ of 0.567 degrees.

\begin{figure}[h]
\center
\includegraphics[scale=0.18]{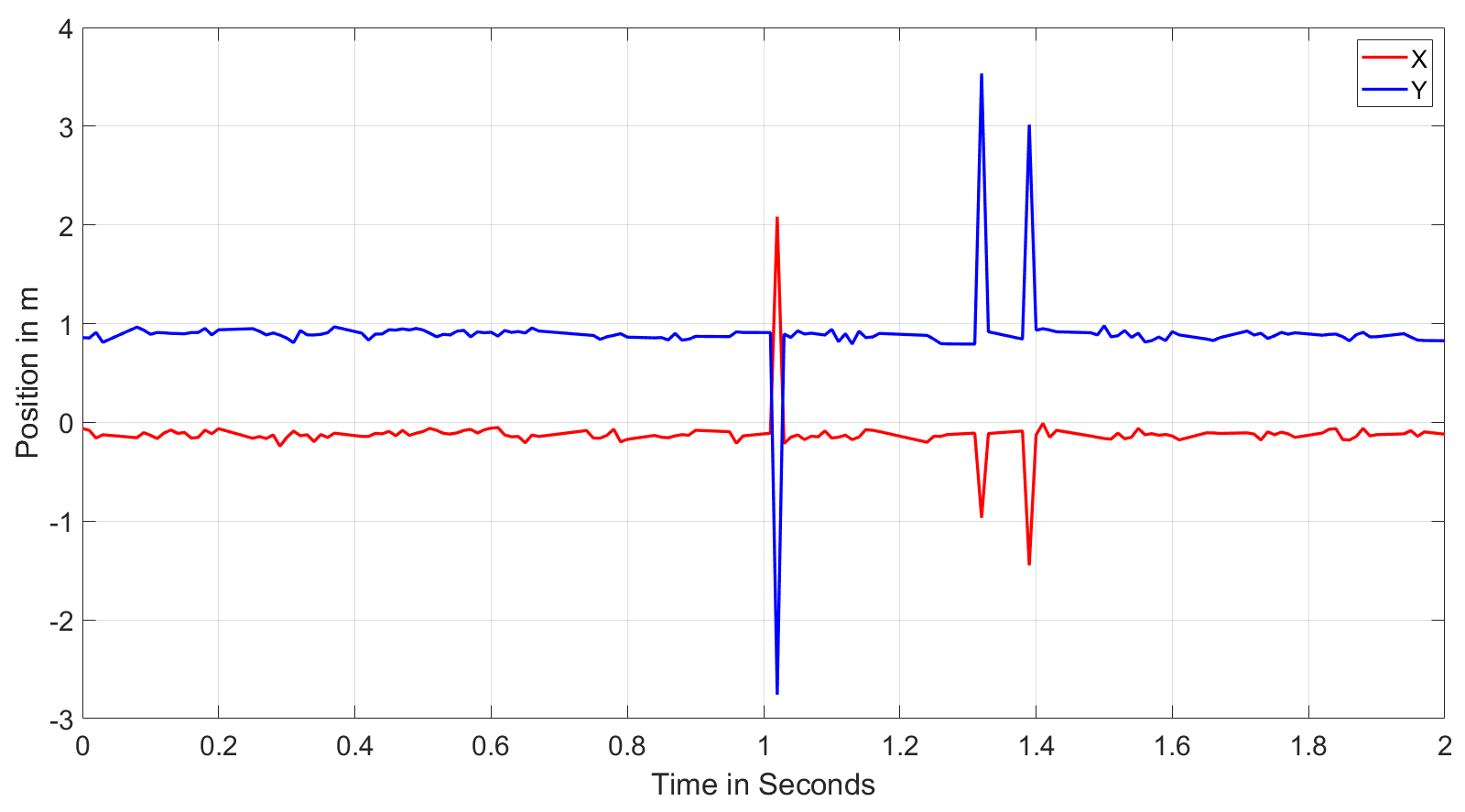}
\caption{Ambiguous solutions for the PnP problem rising from near planar landmarks layout resulting in large positioning errors.}
\label{fig:flips}
\end{figure}

The performance of the TDKF depends on the accuracy of the measurement corrections utilized from solving the PnP problem. If the landmarks were placed in a near planar constellation, multiple solutions to the PnP problem exist leading to undesired flip ambiguities, see Fig. \ref{fig:flips}. To avoid such ambiguities, we have conducted several experiments with various layouts of the landmarks' placement on the UGV. It was concluded that at least two landmarks need to be placed, such that the distance between them along the NVS optical axis is equal to one-seventh the maximum distance between the UAV and UGV. For our test scenario, the maximum distance between the UAV and UGV was 7 m and hence, two landmarks needed to placed 1 m apart along the optical axis of the NVS, as shown in figure \ref{fig:drone_ugv}. 
Given the geometrical constraints of the Perseverance Rover, $d_{l}$ can be at most 3 m. Hence, we conjecture that the relative localization range could increase to approximately 21 m. More importantly, landmarks with higher luminous intensities need to be utilized to ensure that they are detected from such a range. As 1300 mcd landmarks were detected from 7 meter range and since light propagation follows the inverse square law \cite{inverse_square}, the required illumination to trigger the NVS pixels must be 27 mcd. This implies that for landmarks to be detected from 20-meter range, the required illumination needs to be 10,600 mcd. It is fundamentally important to highlight that the measurement corrections for the LTKF are independent of the landmarks layout since they are clustered based on their unique transition frequencies regardless of their positions in the image plane as long as they are not completely occluded.

\begin{table*}[ht]
\centering
\caption{The developed relative localization system is compared to the works of Faessler et al., Chen et al. and Li et al. Our system surpasses the performance of previous SOTA works in estimating the relative pose of the platform within a larger range. The dash indicates that the metric was not mentioned in the article.}
\resizebox{\textwidth}{!}{%
\begin{tabular}{|c|c|c|c|c|c|}
\hline
\textit{\textbf{Authors}} & \textit{\textbf{Breitenmoser et al. \cite{monocular_6d}}} & \textit{\textbf{Faesslar et al. \cite{pose_infrared}}} & \textit{\textbf{Dias et al. \cite{dias}}} & \textit{\textbf{Chen et al. \cite{flicker2}}} & \textit{\textbf{Ours}} \\ \hline
Year & 2011 & 2014 & 2016 & 2020 & \textbf{2022} \\ \hline
Vision Sensor & CMOS & CMOS & CMOS & NVS & \textbf{NVS} \\ \hline
Sensor Resolution & 752 x 480 & 752 x 480 & 752 x 480 & 346 x 260 & \textbf{640 x 480} \\ \hline
\begin{tabular}[c]{@{}c@{}}Maximum \\ Positioning Error (m)\end{tabular} & 0.12 & 0.0328 & 0.25 & 0.03 & \textbf{0.0137} \\ \hline
\begin{tabular}[c]{@{}c@{}}Mean\\ Positioning Error (m)\end{tabular} & 0.015 & 0.0074 & 0.17 & - & \textbf{0.0052} \\ \hline
\begin{tabular}[c]{@{}c@{}}Maximum\\ Orientation Error ($\circ$) \end{tabular} & 4.5 & 3.37 & 12.5 & - & \textbf{2.16} \\ \hline
\begin{tabular}[c]{@{}c@{}}Mean\\ Orientation Error ($\circ$) \end{tabular} & 1.2 & 0.79 & - & - & \textbf{0.567} \\ \hline
Range (m) & 1.75 & 5.5 & 3.5 & 1 & \textbf{7} \\ \hline
Field of View ($\circ$) & 90 & 90 & 90 & 45 & \textbf{45} \\ \hline
\begin{tabular}[c]{@{}c@{}}Pixel\\ Coverage Area (mm${}^{2}$)\end{tabular} & 213.16 & 213.2 & 26.0 & 5.76 & \textbf{88.36} \\ \hline
\begin{tabular}[c]{@{}c@{}}Positioning\\ Percentage Error\end{tabular} & 7\% & 0.13\% & 4.85\% & 3\% & \textbf{0.074\%} \\ \hline
\end{tabular}%
}
\label{table:benchmarks}
\end{table*}

\subsection{Benchmarks}

To evaluate the impact of our developed relative localization system, we compare our results with previous VBM-based relative localization systems from the literature. We benchmark our system performance against the works of Breitnmoser et al. \cite{monocular_6d}, Faessler et al. \cite{pose_infrared}, and Dias et al. \cite{dias} utilizing conventional imaging sensors to observe active LED luminaries for relative localization. We also compare our work against the work of Chen et al. \cite{flicker2} where an NVS is utilized with flickering markers for localization of a single moving platform. Table \ref{table:benchmarks} lists our results compared to those works.

Our approach shows breakthrough performance for both relative position and orientation estimation. Compared to Faessler et al. \cite{pose_infrared}, which provides the best localization accuracy out of all previous works with a higher resolution camera compared to ours, the developed system contributed to a 0.02 m and 1$^\circ$ improvement in positioning and orientation accuracy, respectively, within a longer range and a lower resolution sensor. We also benchmark our system against the approach developed by Chen et al. \cite{flicker2}. Even though we utilized a higher resolution NVS, the relative localization range was 7 times larger than theirs which made it more challenging to position the UAV. Hence, we show that the our sensor pixel coverage area is much larger compared to theirs. We assume that the $FOV$ is $45^{\circ}$ since the same lens is provided for both NVS models. The pixel coverage area is computed by finding the focal length $f$ from the $FOV$

\begin{equation}
    FOV = 2\tan^{-1}\left(\frac{W}{2f}\right),
\end{equation}

\noindent where $W$ is the width of the image plane and then the pixel area from the normalized image coordinates is computed given the range. Accordingly, the covered pixel area computed from the perspective projection model for our scenario tends to be 88.36 mm$^{2}$ much larger than in their approach, 5.76 mm$^{2}$.

On the other hand, the average system execution time is also reported for each system block, see Table \ref{table:exec_time}. The proposed approach shows real time applicability with a total execution time of 2.933 $ms$ when compared to Faesslar et al. \cite{pose_infrared} whose total execution time is 3.8 ms. It is worth noting that the Faesslar et al. \cite{pose_infrared} approach was executed on C++ while ours was developed using Matlab on Intel i7-10750H (2.60GHz) processor. Thus, our execution time can be further improved if implemented using C++.

\begin{table}[h]
\centering
\caption{Execution times for the system blocks.}
\begin{tabular}{|c|c|}
\hline
\textit{\textbf{Algorithm}} & \textit{\textbf{Execution Time (ms)}} \\ \hline
\begin{tabular}[c]{@{}c@{}}Polarity Transition\\Events Construction\end{tabular} & \textit{0.003} \\ \hline
\begin{tabular}[c]{@{}c@{}}Unsupervised Clustering\\ of Polarity Transition Events\end{tabular} & \textit{0.12} \\ \hline
\begin{tabular}[c]{@{}c@{}}LTKF Prediction\end{tabular} & \textit{0.47} \\ \hline
Madgwick Filter & \textit{1.13} \\ \hline
PnP Pose Estimation & \textit{0.97} \\ \hline
TDKF Prediction & \textit{0.24} \\ \hline
\textbf{Total} & \textbf{2.933} \\ \hline
\end{tabular}%
\label{table:exec_time}
\end{table}

\section{Conclusion}
\label{section:conc}
In this paper, a robust relative localization system based on NVBMs aided with flickering landmarks has been developed and was addressed with an application for the Ingenuity and Perseverance Rover. Flickering landmarks were adopted on the targeted agents to guarantee their continuous observability by the neuromorphic vision sensor onboard the aerial vehicle, even in absence of relative motion between both robots. This was attained by developing a novel unsupervised clustering approach for the events for landmark identification and by designing state estimators, LTKF and TDKF, for robust tracking and relative localization without relying on conventional imaging techniques. The system was validated in experiments resembling the targeted space scenario where the proposed approach outperformed state-of-the-art methods in terms of localization accuracy and execution time. Even though the developed relative localization system was deployed in a space application, its robust performance opens up future opportunities for deployment in a wide variety of applications, such as relative localization of robotic swarms in space, localization systems for indoor and underwater positioning systems due to the unavailability of GPS in such areas. Future work will extend the system to those applications and utilize motion models for observed agents in robotic swarms. In addition, the influence of the landmarks arrangement will be investigated for defining their optimal layout to further improve the system localization accuracy.

\section*{Acknowledgment}
The authors gratefully acknowledge the efforts of Osama Abdulhay in the experiments.

\bibliographystyle{IEEEtran}
\bibliography{main.bib}

\begin{IEEEbiography}[{\includegraphics[width=1in,height=1.25in,clip,keepaspectratio]{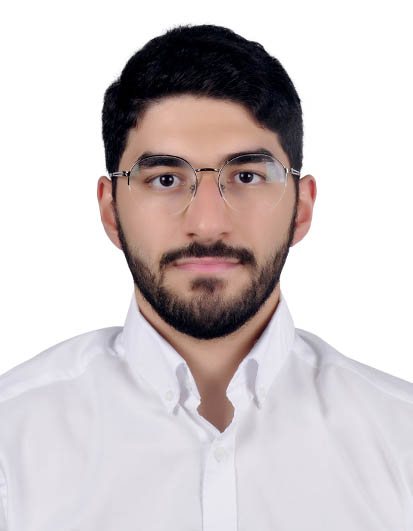}}]%
{Mohammed Salah}
received his BSc. in Mechanical Engineering from American University of Sharjah, UAE, in 2020 and his MSc. in Mechanical Engineering from Khalifa University, Abu Dhabi, UAE, in 2022. He is currently with Khalifa University Center for Autonomous Robotic Systems (KUCARS). His research interest is mainly focused on multisensor fusion, neuromorphic vision, and space robotics.
\end{IEEEbiography}

\begin{IEEEbiography}[{\includegraphics[width=1in,height=1.25in,clip,keepaspectratio]{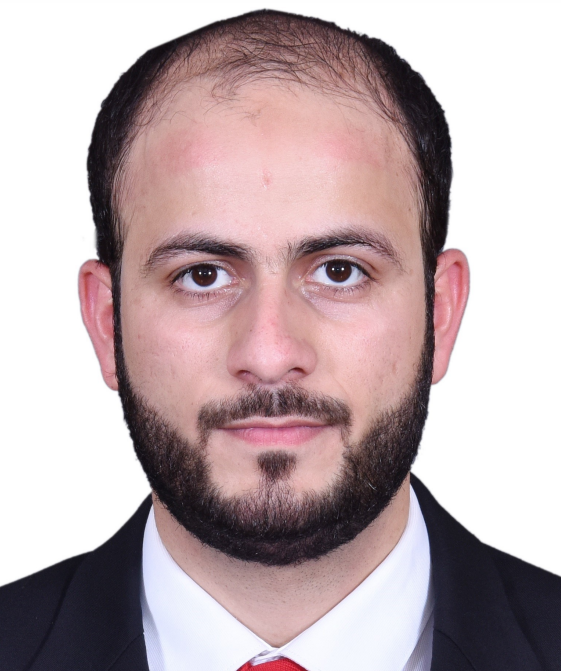}}]%
{Mohammed Chehadah}
(Member, IEEE) received his MSc. in Electrical Engineering from Khalifa University, Abu Dhabi, UAE, in 2017. He is currently with Khalifa University Center for Autonomous Robotic Systems (KUCARS). His research interest is mainly focused on identification, perception, and control of complex dynamical systems utilizing the recent advancements in the field of AI.
\end{IEEEbiography}

\begin{IEEEbiography}[{\includegraphics[width=1in,height=1.25in,clip,keepaspectratio]{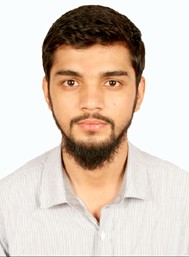}}]%
{Muhammad Humais}
received his M.Sc. in Electrical and Computer Engineering from Khalifa University in 2020. His research is mainly focused on robotic perception and control for autonomous systems. He is currently a Ph.D. fellow at Khalifa University Center for Autonomous Robotics (KUCARS). 
\end{IEEEbiography}

\begin{IEEEbiography}[{\includegraphics[width=1in,height=1.25in,clip,keepaspectratio]{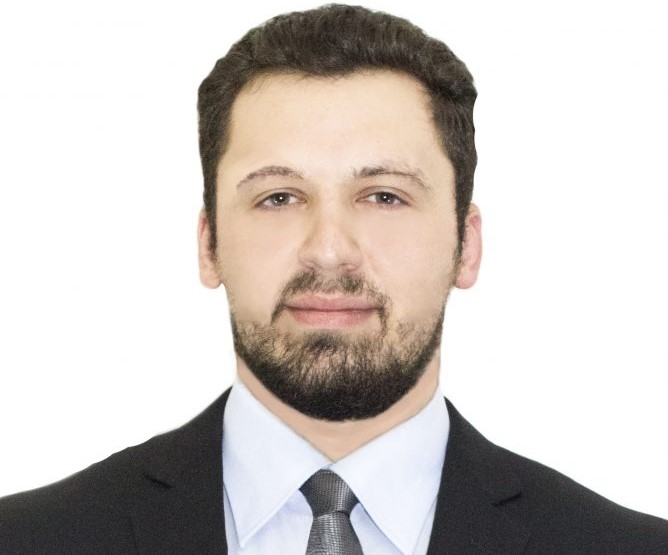}}]%
{Mohammed Wahbah}
received his MSc. in Electrical Engineering from Khalifa University, Abu Dhabi, UAE, 2018. He is currently a researcher with Khalifa University Center for Autonomous Robotic Systems (KUCARS). His research areas include multisensor fusion, state estimation, and navigation in hazardous environments.
\end{IEEEbiography}

\begin{IEEEbiography}[{\includegraphics[width=1in,height=1.25in,clip,keepaspectratio]{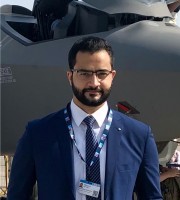}}]%
{Abdulla Ayyad}
(Member, IEEE) received the M.Sc. degree in electrical engineering from The University of Tokyo, in 2019, where he conducted research with the Spacecraft Control and Robotics Laboratory. He is currently a Research Associate at the Advanced Research and Innovation Center (ARIC) at Khalifa University working on several robot autonomy projects. His current research interest includes the application of AI in the fields of perception, navigation, and control.
\end{IEEEbiography}

\begin{IEEEbiography}[{\includegraphics[width=1in,height=1.25in,clip,keepaspectratio]{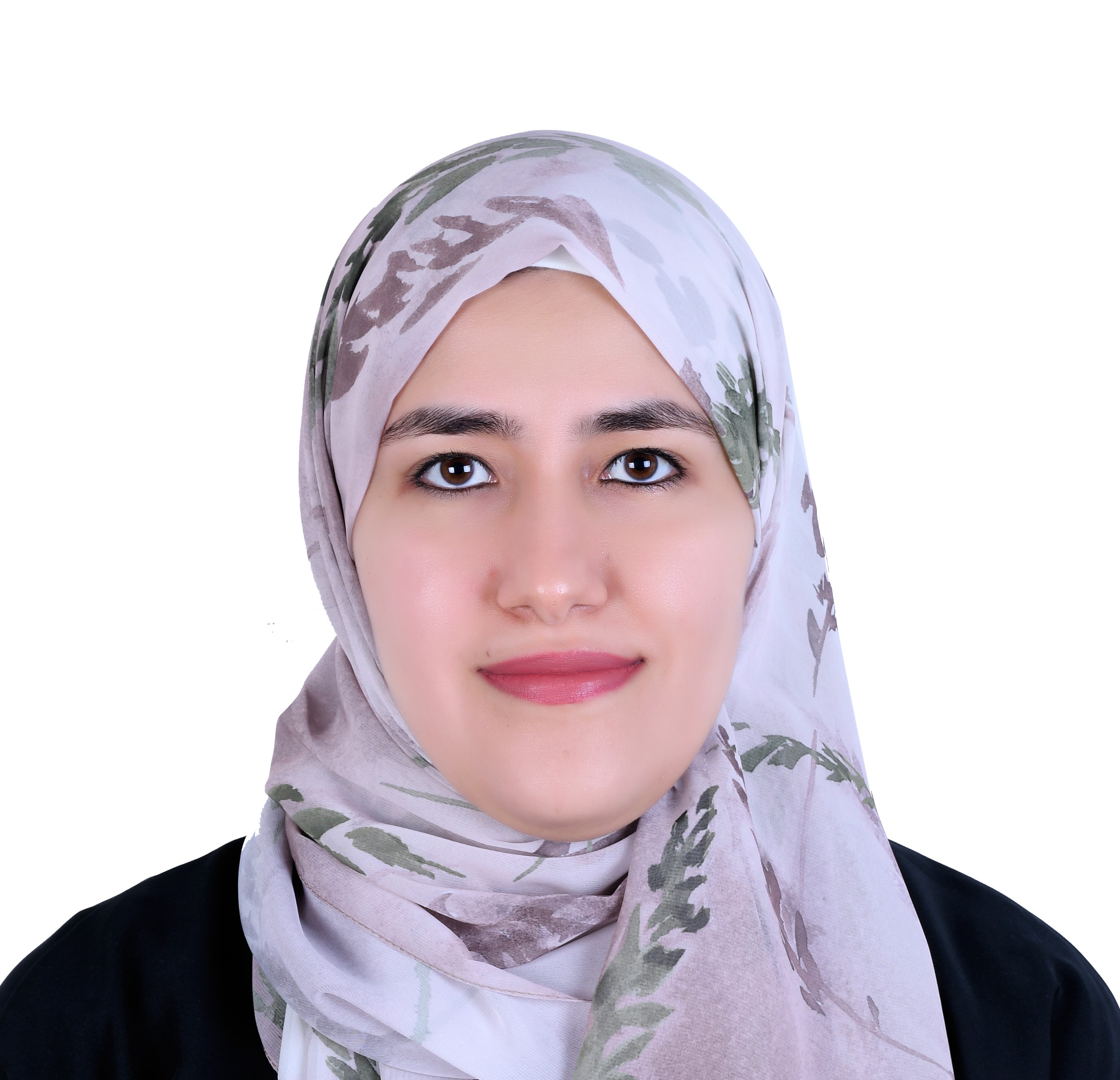}}]%
{Rana Azzam} received the B.Sc. degree in computer engineering, the M.Sc. degree by Research in electrical and computer engineering, and the Ph.D. degree in engineering with a focus on robotics from Khalifa University in 2014, 2016, and 2020 respectively. She is currently a Postdoctoral Fellow with the Department of Aerospace Engineering. Her research interests include machine learning, reinforcement learning, navigation, and simultaneous localization and mapping.
\end{IEEEbiography}

\begin{IEEEbiography}[{\includegraphics[width=1in,height=1.25in,clip,keepaspectratio]{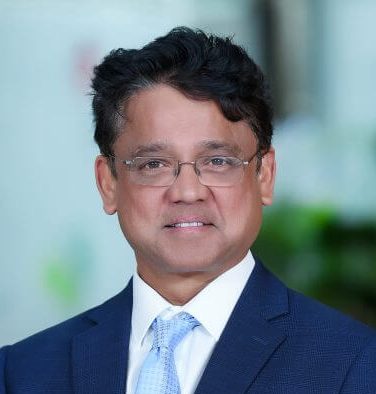}}]%
{Lakmal Seneviratne}
is Professor of Mechanical Engineering and the founding Director of the Centre for Autonomous Robotic Systems (KUCARS) at Khalifa University, UAE. He has also served as Associate Provost for Research and Graduate Studies and Associate VP Research at Khalifa University. Prior to joining Khalifa University, he was Professor of Mechatronics, the founding Director of the Centre for Robotics Research and the Head of the Division of Engineering, at King’s College London. He is Professor Emeritus at King’s College London. His main research interests are centered on robotics and automation, with particular emphasis on increasing the autonomy of robotic systems interacting with complex dynamic environments. He has published over 400 peer reviewed publications on these topics. He is a member of the Mohammed Bin Rashid Academy of Scientists in the UAE.
\end{IEEEbiography}

\begin{IEEEbiography}[{\includegraphics[width=1in,height=1.25in,clip,keepaspectratio]{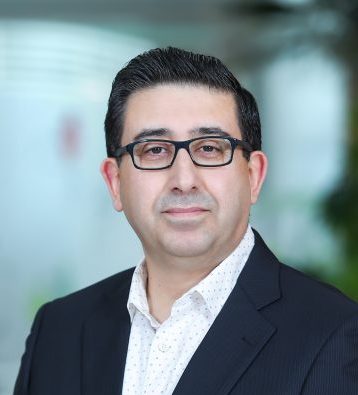}}]%
{Yahya Zweiri} (Member, IEEE) received the Ph.D. degree from the King’s College London in 2003. He is currently an Associate Professor with the Department of Aerospace Engineering and deputy director of Advanced Research and Innovation Center - Khalifa University, United Arab Emirates. He was involved in defense and security research projects in the last 20 years at the Defense Science and Technology Laboratory, King’s College London, and the King Abdullah II Design and Development Bureau, Jordan. He has published over 130 refereed journals and conference papers and filed ten patents in USA and U.K., in the unmanned systems field. His main expertise and research are in the area of robotic systems for extreme conditions with particular emphasis on applied Artificial Intelligence (AI) aspects and neuromorphic vision system.
\end{IEEEbiography}

\end{document}